\documentclass[runningheads]{llncs}
\usepackage[T1]{fontenc}
\usepackage{amsfonts}
\usepackage{amsmath}
\usepackage[hyphens]{url}
\usepackage{siunitx}
\usepackage{graphicx}
\usepackage{booktabs}
\usepackage{multirow}
\usepackage[table]{xcolor}
\usepackage{colortbl} 
\usepackage{xcolor}   
\begin{document}
\title{CoMViT: An Efficient Vision Backbone for Supervised Classification in Medical Imaging}
\titlerunning{CoMViT: Efficient Vision Backbone for Medical Imaging}
%

\author{Aon Safdar\inst{1}\orcidID{0000-0003-3039-1017} \and
Mohamed Saadeldin\inst{1}\orcidID{0000-0003-4104-9524}}
\authorrunning{A. Safdar and M. Saadeldin}
%
\institute{School of Computer Science, University College Dublin, Republic of Ireland \\
\email{aon.safdar@ucdconnect.ie}, \email{mohamed.saadeldin@ucd.ie}}
\maketitle              

\begin{abstract}
Vision Transformers (ViTs) have demonstrated strong potential in medical imaging; however, their high computational demands and tendency to overfit on small datasets limit their applicability in real-world clinical scenarios. In this paper, we present \textbf{CoMViT}\footnote{Code available at: \url{https://github.com/aonsafdar/CoMViT}}, a compact and generalizable Vision Transformer architecture specifically optimized for resource-constrained medical image analysis. CoMViT integrates a convolutional tokenizer, diagonal masking, dynamic temperature scaling, and pooling-based sequence aggregation to improve performance and generalization. Through systematic architectural optimization, CoMViT achieves robust performance across twelve MedMNIST datasets while maintaining a lightweight design with only $\sim$4.5M parameters. It matches or outperforms deeper CNN and ViT variants, offering up to $5$--$20\times$ parameter reduction without sacrificing accuracy. Qualitative Grad-CAM analyses further reveal that CoMViT consistently attends to clinically relevant regions despite its compact size. Our findings highlight the potential of principled ViT re-design for developing efficient and interpretable models in low-resource medical imaging settings.

\end{abstract}

\section{Introduction}
\label{sec:intro}

Vision Transformers (ViTs) have emerged as powerful vision backbones due to their ability to model long-range dependencies via self-attention~\cite{dosovitskiy_image_2021,touvron_training_2021}. Their scalability has led to widespread adoption across classification, detection, and segmentation tasks~\cite{han_survey_2023}. However, their use in medical imaging remains limited, particularly in low-resource settings, where challenges such as small dataset sizes, scarce annotations, domain heterogeneity, and constrained compute persist~\cite{aburass_vision_2025,li_transforming_2023}.

Several strategies aim to mitigate ViTs’ data inefficiency. Hybrid CNN-ViT models introduce inductive biases to preserve spatial locality~\cite{liu_feature_2022,manzari_medvit_2023}, though at the cost of added architectural complexity. Others pursue transfer learning by distilling CNNs~\cite{touvron_training_2021} or fine-tuning large ViTs pretrained on natural images~\cite{halder_implementing_2024}. These often struggle with domain shift, leading to sub-optimal transfer or negative knowledge transfer (NKT)~\cite{du_mdvit_2023}.

Efforts to scale data—via augmentation~\cite{avidan_towards_2022}, unsupervised pretraining~\cite{caron_emerging_2021}, or multi-domain training—offer partial remedies but struggle with generalizability. Training across diverse domains may help one dataset while degrading performance on others. Domain-adaptive methods using CNN-based adapters~\cite{du_mdvit_2023} reduce this effect but add parameter overhead proportional to the number of domains.

We take a different approach. Rather than relying on scale, transfer, or domain-specific adaptation, we introduce \textbf{CoMViT}—a lean and universal ViT backbone for data- and compute-constrained medical imaging. CoMViT integrates several synergistic architectural innovations to boost efficiency and representational power. Specifically, (1) A shallow convolutional tokenizer that encodes local context early, adding strong spatial priors. (2) A lightweight transformer encoder with empirically chosen depth, embedding size, and MLP/head widths~\cite{compact}, balancing accuracy and efficiency. (3) Diagonal masking and learnable temperature scaling~\cite{lee_improving_2022} to promote localized attention and gradient stability. (4) Sequence pooling in place of the classification token to improve aggregation and reduce redundancy.

We evaluate CoMViT across all twelve 2D MedMNIST datasets spanning multiple modalities (e.g., X-ray, OCT, microscopy) and diagnostic tasks. CoMViT consistently matches or exceeds the performance of significantly larger CNN and ViT baselines, using $5\times$–$20\times$ fewer parameters and FLOPs. Qualitative results confirm its focus on disease-relevant regions despite its compactness.

Our contribution are as under:
\begin{itemize}
    \item We propose a compact ViT backbone optimized for data- and compute-limited medical imaging scenarios.
    \item We demonstrate that lightweight transformers can outperform deeper models across diverse medical modalities with superior accuracy-efficiency tradeoffs.
    \item We provide qualitative insights showing interpretable and diagnostically relevant attention maps.
\end{itemize}

Our results show that scale is not a prerequisite for success in medical imaging. Thoughtful architectural design can yield compact, generalizable ViTs suitable for real-world, resource-constrained deployment.


\section{Related Work}

The MedMNIST benchmark~\cite{yang_medmnist_2023} introduced a suite of 2D biomedical datasets with low-resolution images across diverse modalities, enabling efficient model prototyping. Early baselines leveraged convolutional neural networks (CNNs) for their strong spatial priors, but CNNs require deep stacks to model long-range dependencies, increasing compute and memory cost~\cite{aburass_vision_2025}.

Recent efforts have adapted ViTs for medical imaging. FPViT~\cite{liu_feature_2022} integrates ResNet feature pyramids as token inputs to enhance patch encoding. MedViT~\cite{manzari_medvit_2023} introduces local convolutional attention and depthwise MLPs for efficient context modeling. These methods improve locality but add architectural complexity. Fine-tuning large pretrained ViTs~\cite{halder_implementing_2024} yields competitive results but often suffers from domain mismatch and negative knowledge transfer~\cite{du_mdvit_2023}. Other works explore ViT robustness and domain alignment. PyramidAT~\cite{herrmann_cvpr_nodate} enhances resilience via multi-scale adversarial training; SelfCSL~\cite{nguyen_semi-supervised_2021} leverages contrastive learning for semi-supervised domain adaptation; MedRDF~\cite{xu_medrdf_2022} uses voting-based ensembling to defend against adversarial attacks.
Despite these advances, lightweight ViT backbones tailored for low-resource medical imaging remain scarce. \textbf{CoMViT} addresses this gap by achieving competitive accuracy with $\sim$4.5M parameters, surpassing deeper CNNs and larger ViTs—while maintaining practical efficiency for deployment in constrained clinical settings.

\section{Method}
\label{sec:method}

\begin{figure}[t]
\centering
\includegraphics[width=\linewidth]{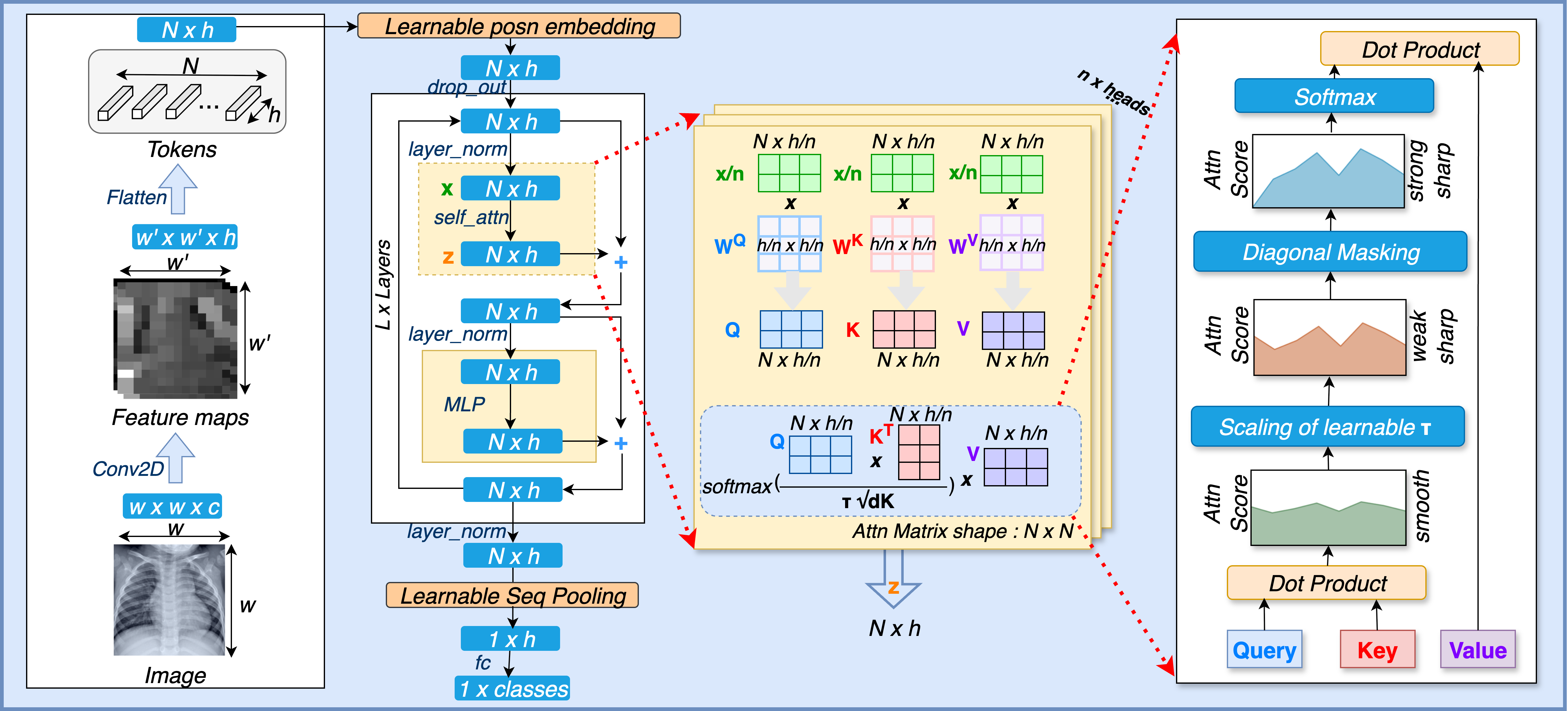}
\caption{\textbf{CoMViT Architecture Overview}}
\label{fig:architecture}
\end{figure}

We propose \textbf{ComViT}, a compact Vision Transformer designed to address the limitations of standard ViTs in resource-constrained medical imaging tasks. ComViT departs from rigid patch-based tokenization and heavy attention mechanisms, offering a streamlined design optimized for performance, efficiency, and interpretability. Fig.~\ref{fig:architecture} outlines the end-to-end model pipeline.

Given an input image $x \in \mathbb{R}^{w \times w \times c}$, ComViT first employs a lightweight convolutional stem with two layers to generate spatial feature maps. Instead of the rigid patch-splitting used in ViT, we utilize a shallow convolutional tokenizer. This tokenizer replaces fixed-size patches with learned local filters, promoting inductive bias and better spatial encoding. Specifically, it consists of two $7 \times 7$ Conv2D layers followed by a $3 \times 3$ max-pooling layer. The resulting feature maps are flattened into a sequence of $N$ tokens, each $h$-dimensional, yielding $T \in \mathbb{R}^{N \times h}$. Learnable positional embeddings $P \in \mathbb{R}^{N \times h}$ are added to preserve spatial structure, forming $Z_0 = T + P$.

The sequence is then processed by $L$ stacked Transformer layers. Each layer employs multi-head self-attention with enhancements:

\begin{equation}
\begin{aligned}
A_{ij} &= \frac{Q_i K_j^\top}{\tau \sqrt{d_k}} + M, \\
\text{head}_k &= \text{softmax}(A)V, \\
Z &= \text{Concat}(\text{head}_1, \ldots, \text{head}_n)W^O
\end{aligned}
\end{equation}

Here, $\tau$ is a learnable temperature~\cite{lee_improving_2022}, and $M$ is a diagonal mask with $M_{ii} = -\infty$ to suppress self-attention~\cite{ferdous_spt-swin_2024}. Each block is followed by a position-wise MLP with GELU activation and wrapped in residual connections with layer normalization.

We adopt \emph{learnable sequence pooling} for final representation aggregation, replacing the static class token. This allows the network to adaptively weight tokens based on relevance. This approach improves efficiency and avoids the overhead of an explicit global token.

\begin{equation}
s = \sum_{i=1}^N w_i O^i, \quad w_i = \text{softmax}(W_p O^i)
\end{equation}
As shown in Table \ref{tab:model-config}, we configure ComViT with 7 layers, a 256-dim hidden size, 4 attention heads, and an MLP expansion of 2×, totaling only 4.5M parameters. These settings, inspired by~\cite{hassani_escaping_2022} and further optimized, balance compactness and performance. Through this combination of learned tokenization, attention refinement, and adaptive pooling, ComViT delivers an efficient and interpretable ViT backbone suited for low-resource medical imaging.
\section{Experiments and Results}

\subsection{Datasets}

\begin{figure}[ht]
\centering
\includegraphics[width=0.95\textwidth]{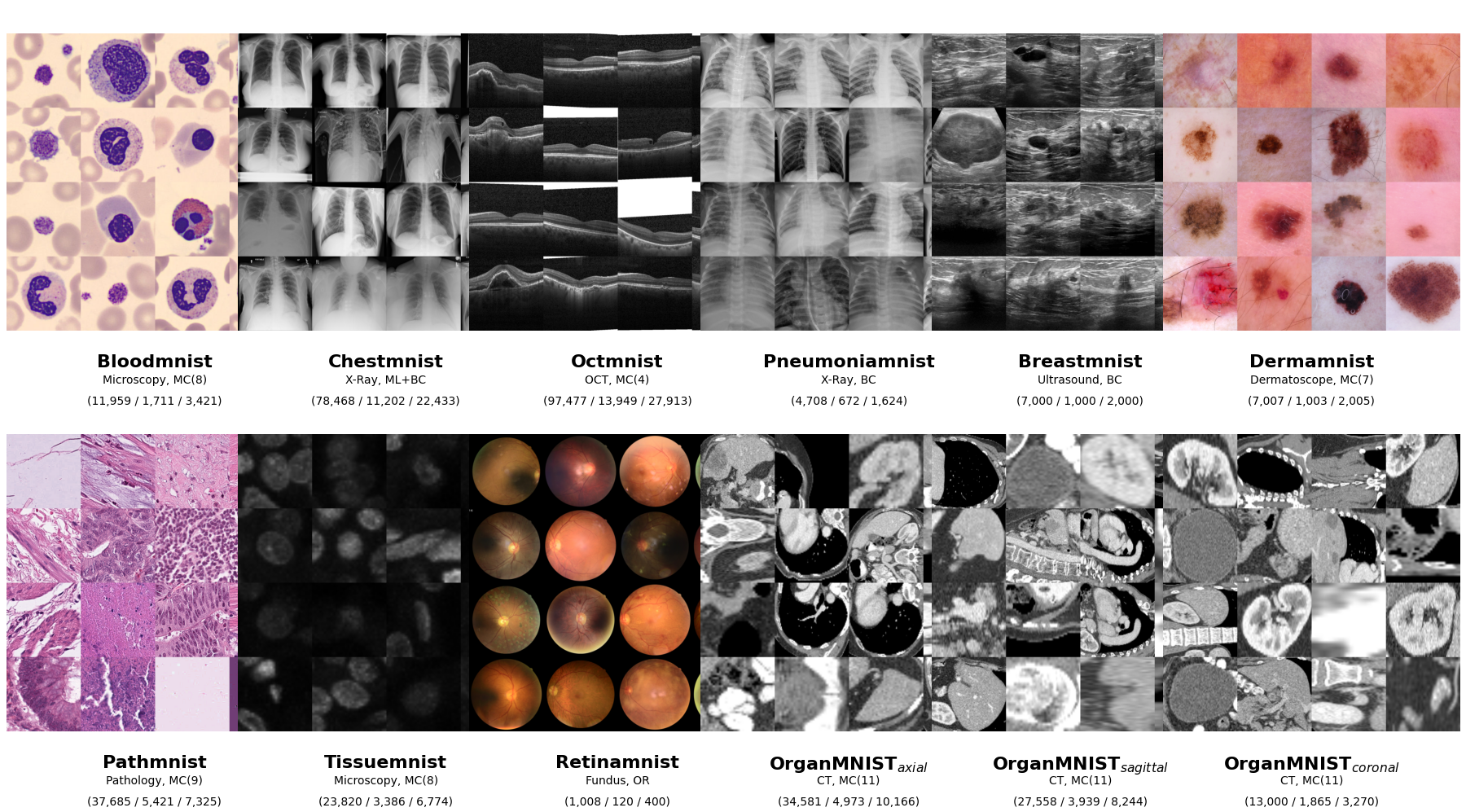}
\caption{Overview of the MedMNIST Dataset Family. Modality, Classification Task, and split (Train / Val / Test) are mentioned. Best viewed in color and zoomed in.}
\label{fig:medmnist}
\end{figure}

\begin{table}[t]
\centering
\caption{
\textbf{ViT model configurations.} ComViT adopts an empirically tuned lightweight setup optimized for medical imaging.
}
\vspace{1mm}
\begin{tabular}{lcccccc}
\toprule
\textbf{Model} & \textbf{Layers} & \textbf{Hidden Size} $D$ & \textbf{MLP Size} & \textbf{Heads} & \textbf{Params} & \textbf{Source} \\
\midrule
ViT-H       & 32  & 1280 & 5120  & 16 & 632M  & \cite{dosovitskiy_image_2021} \\
ViT-L       & 24  & 1024 & 4096  & 16 & 307M  & \cite{dosovitskiy_image_2021,steiner_how_2022} \\
ViT-B       & 12  & 768  & 3072  & 12 & 86M   & \cite{dosovitskiy_image_2021,steiner_how_2022} \\
ViT-S       & 12  & 384  & 1536  & 6  & 22.1M & \cite{steiner_how_2022} \\
ViT-Ti      & 12  & 192  & 768   & 3  & 5.8M  & \cite{steiner_how_2022} \\
\textbf{ComViT} & \textbf{7} & \textbf{256} & \textbf{512} & \textbf{4} & \textbf{4.5M} & Ours \\
\bottomrule
\end{tabular}
\label{tab:model-config}
\end{table}
MedMNIST~\cite{yang_medmnist_2023} is a curated suite of 12 biomedical imaging datasets covering diverse modalities such as X-ray, histopathology, dermatoscope, and microscopy (Fig.~\ref{fig:medmnist}). All datasets are standardized to 224$\times$224 resolution, enabling reproducibility. Unlike task-specific datasets, MedMNIST provides a modality-agnostic benchmark for evaluating generalizable tokenization and attention strategies. It supports multi-class, binary, and ordinal tasks, spanning texture- and structure-rich modalities under a unified evaluation protocol.

\subsection{Model Configuration}

ComViT uses a shallow tokenizer with two convolution and max-pooling layers to produce overlapping local tokens. A 7-layer transformer encoder follows, with 4 attention heads and a hidden size of 256. Attention is enhanced using diagonal masking (self-attention suppression) and learnable temperature scaling~\cite{lee_improving_2022}. We use a $2{\times}$ expansion MLP per block. Final representations are aggregated via learnable sequence pooling~\cite{compact}, replacing the CLS token.

\subsection{Experiment Protocol}

\paragraph{Hyperparameters.}
ComViT is trained in PyTorch with \texttt{timm}~\cite{Wightman_PyTorch_Image_Models}, using AdamW for 300 epochs. The learning rate starts at $1.1 \times 10^{-4}$ with cosine decay, warm-up (10 epochs), and cooldown (10 epochs). Regularization includes RandAugment, Mixup ($\alpha=0.8$), CutMix ($\alpha=1.0$), label smoothing, and drop-path (0.1). Mixup is probabilistically turned off after epoch 175. AMP and gradient clipping (max norm 1.0) are enabled. Batch size is 512 and input size is 224$\times$224.

\paragraph{Benchmarks and Metrics.}
We compare against strong ViT baselines (DeiT-Ti, PiT-Ti, PVT-Ti, RVT-Ti) and CNNs (ResNet-18, EfficientNet-B3) from~\cite{manzari_medvit_2023}. Evaluation includes Top-1 test accuracy, parameter count, and GFLOPs per forward pass—capturing tradeoffs in accuracy, memory, and compute.

\begin{table}[t]
\centering
\caption{Accuracy (\%) comparison across MedMNIST2D datasets with 224$\times$224 resolution. \textcolor{red}{Red} is best, \textcolor{blue}{blue} is second-best.}
\label{tab:datasetwise_acc}
\resizebox{\textwidth}{!}{%
\begin{tabular}{lcccccccccccc}
\toprule
\textbf{Method} & Path & Chest & Derma & OCT & Pneumonia & Retina & Breast & Blood & Tissue & OrganA & OrganC & OrganS \\
\midrule
ResNet-18        & 90.9 & 94.7 & 75.4 & 76.3 & 86.4 & 49.3 & 83.3 & \textcolor{blue}{96.3} & 68.1 & \textcolor{blue}{93.6} & \textcolor{blue}{92.0} & 78.5 \\
ResNet-50        & \textcolor{blue}{92.0} & 94.8 & \textcolor{blue}{77.3} & 76.2 & 88.4 & 51.1 & \textcolor{blue}{84.2} & 96.0 & 68.0 & 93.5 & 91.1 & 77.0 \\
auto-sklearn     & 71.6 & 77.9 & 71.9 & 60.1 & 85.5 & 51.5 & 80.3 & 87.8 & 53.2 & 76.2 & 82.9 & 67.2 \\
Google AutoML    & 72.8 & 77.8 & \textcolor{red}{77.8} & \textcolor{blue}{76.8} & 91.6 & 53.1 & 86.1 & 90.8 & 67.3 & 88.6 & 87.7 & 74.9 \\
MedViT-Tiny      & \textcolor{red}{95.6} & \textcolor{red}{95.6} & 76.8 & 76.7 & \textcolor{red}{94.9} & \textcolor{blue}{53.4} & \textcolor{red}{89.6} & 95.0 & \textcolor{red}{70.3} & 93.1 & 90.1 & \textcolor{blue}{78.9} \\
\textbf{CoMViT}  & 91.08 & \textcolor{blue}{95.12} & 77.0 & \textcolor{red}{84.6} & \textcolor{blue}{92.14} & \textcolor{red}{53.9} & 83.97 & \textcolor{red}{98.04} & \textcolor{blue}{69.79} & \textcolor{red}{95.15} & \textcolor{red}{92.84} & \textcolor{red}{80.44} \\
\bottomrule
\end{tabular}%
}
\end{table}
\section{Results and Discussion}

We evaluate CoMViT across all 12 datasets from MedMNIST2D and benchmark it against widely-used CNNs (ResNet-18/50), AutoML systems (AutoKeras, Google AutoML, auto-sklearn), and Transformer variants (MedViT-T). Table~\ref{tab:datasetwise_acc} shows that CoMViT achieves the best or second-best accuracy on 8 datasets, matching or outperforming much larger models. This highlights CoMViT’s strong generalizability across diverse tasks and imaging modalities.
\begin{figure}[htbp]
\centering
\includegraphics[width=\textwidth]{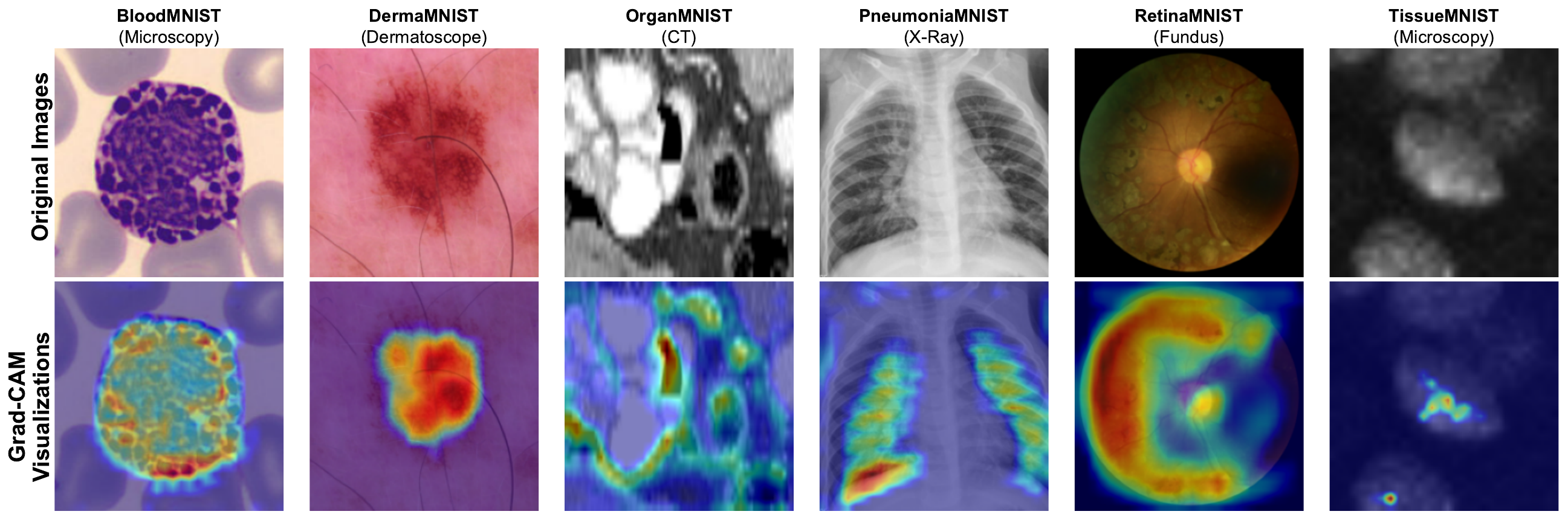}
\caption{Grad-CAM visualizations across six MedMNIST datasets highlighting the regions of interest that contribute most to the model's predictions. CoMViT accurately localizes relevant regions across diverse modalities.}
\label{fig:gradcam_camvit}
\end{figure}

Despite a compact footprint, CoMViT remains competitive with SOTA models. Table~\ref{tab:avg_acc} shows average accuracy and parameter count. CoMViT achieves 84.5\% accuracy using only 4.55M parameters, outperforming ResNet-50 (82.1\%) and matching MedViT-T (84.0\%) which has over 2× parameters.Table~\ref{tab:comparison} compares Tiny/Small/Large models on TissueMNIST. CoMViT leads among Tiny models (69.8\%) while being the lightest. It rivals Small models like Swin-T and Twins-SVT-S with lower complexity, validating its architecture.

\begin{table}[t]
\centering
\caption{Average accuracy comparison on MedMNIST-2D.}
\label{tab:avg_acc}
\scriptsize
\begin{tabular}{lcc}
\toprule
\textbf{Methods} & \textbf{Params (M)} & \textbf{Avg. Top-1 Acc} \\
\midrule
ResNet-18 (224)   & 11.2 & 0.821 \\
ResNet-50 (224)   & 23.5 & 0.821 \\
auto-sklearn      & --   & 0.722 \\
AutoKeras         & --   & 0.813 \\
Google AutoML     & --   & 0.809 \\
MedViT-T (224)    & 10.2 & 0.840 \\
MedViT-S (224)    & 23.0 & 0.851 \\
MedViT-L (224)    & 45.0 & 0.842 \\
\midrule
\rowcolor{gray!10} \textbf{CoMViT (Ours)} & \textbf{4.55} & \textbf{0.845} \\
\bottomrule
\end{tabular}
\end{table}

\begin{table*}[ht]
\centering
\caption{Comparison with Tiny/Small/Large models on TissueMNIST.}
\label{tab:comparison}
\scriptsize
\begin{tabular}{llcccc}
\toprule
\textbf{Segment} & \textbf{Model} & \textbf{Img Size} & \textbf{Params (M)} & \textbf{FLOPs (G)} & \textbf{Top-1 (\%)} \\
\midrule
\multirow{6}{*}{\textbf{Tiny}} 
& ResNet-18  & 224 & 11.7 & 1.8 & 68.1 \\
& DeiT-Ti    & 224 & 5.7  & 1.3 & 59.5 \\
& PiT-Ti     & 224 & 4.9  & 0.7 & 62.1 \\
& PVT-T      & 224 & 13.2 & 1.9 & 63.4 \\
& RVT-Ti     & 224 & 8.6  & 1.3 & 69.6 \\
& \textbf{CoMViT} & 224 & \textbf{4.55} & \textbf{1.6} & \textbf{69.8} \\
\midrule
\multirow{5}{*}{\textbf{Small}} 
& ResNet-50     & 224 & 25.6 & 4.1 & 68.0 \\
& DeiT-S        & 224 & 22.0 & 4.6 & 67.0 \\
& Swin-T        & 224 & 29.0 & 4.5 & 71.7 \\
& Twins-SVT-S   & 224 & 24.0 & 2.9 & 72.1 \\
& MedViT-S      & 224 & 23.6 & 4.9 & 73.1 \\
\midrule
\multirow{4}{*}{\textbf{Large}} 
& ResNet-152    & 224 & 60.2 & 11.3 & 67.5 \\
& DeiT-B        & 224 & 87.0 & 17.5 & 66.9 \\
& Swin-B        & 224 & 87.8 & 15.4 & 68.5 \\
& MedViT-L      & 224 & 45.8 & 13.4 & 69.9 \\
\bottomrule
\end{tabular}
\end{table*}

\begin{figure}[t]
\centering
\includegraphics[width=0.9\linewidth]{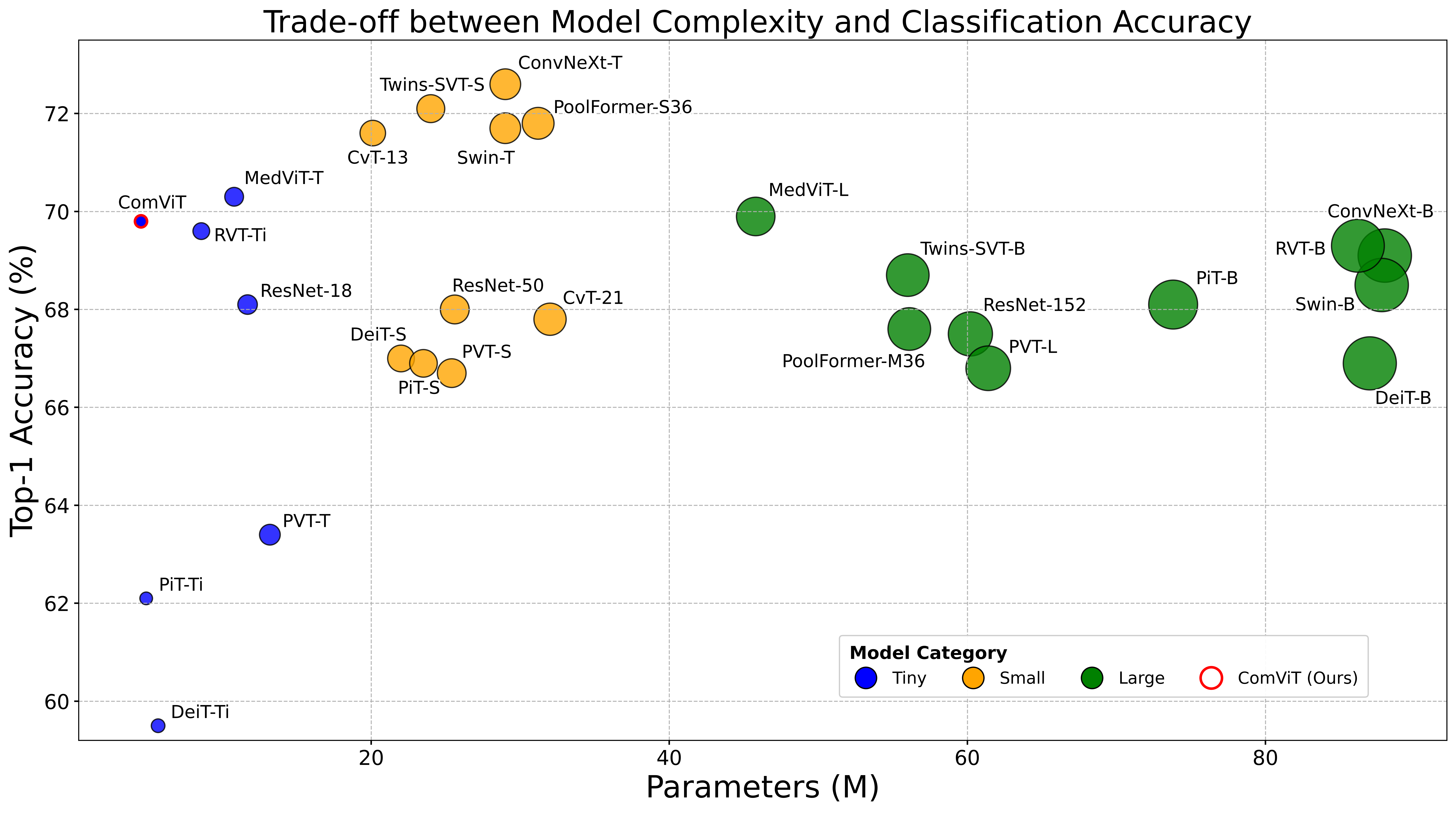}
\caption{\textbf{Model Size vs. Accuracy on TissueMNIST.} CoMViT achieves strong accuracy with minimal parameters, illustrating excellent efficiency.}
\label{fig:size_vs_acc}
\end{figure}

The parameter efficiency and accuracy trade-off are further visualized in Fig.~\ref{fig:size_vs_acc}. CoMViT appears near the Pareto frontier, striking a balance where further increases in model size yield only marginal accuracy gains. In contrast to other methods that require large parameter budgets to generalize well, CoMViT remains scalable and suitable for deployment across a variety of medical imaging tasks—without retraining or architecture tuning. Fig.~\ref{fig:gradcam_camvit} presents Grad-CAM visualizations to assess the interpretability of CoMViT. The model consistently highlights pathology-relevant regions across diverse modalities—such as cell boundaries in BloodMNIST, lesion areas in DermaMNIST, and lung fields in PneumoniaMNIST—indicating that its predictions are grounded in clinically meaningful structures. Notably, in datasets with subtle or diffuse features (e.g., TissueMNIST), CaMViT still localizes relevant regions without relying on spurious artifacts. These results qualitatively support the model’s capacity to extract semantically informative features while preserving spatial and structural priors.

In summary, CoMViT sets a strong baseline for low-resource medical image classification, offering competitive accuracy, lower inference cost, and improved scalability—making it a practical and robust choice for real-world medical AI systems. ComViT’s improvements stem from a synergy of local bias, attention refinement, and compact architecture. The convolutional tokenizer extracts fine-grained local patterns absent in raw patch-based ViTs. Diagnol masking and dynamic temperature scaling complement this by enforcing spatial locality inside the attention heads, limiting unnecessary global mixing, and promoting robust local feature aggregation. Learnable pooling further allows flexible summarization of the token sequence without needing a rigid [CLS] vector, improving adaptation to lesions of varying shapes and locations. The lightweight encoder, moderate embedding size, and learnable positional priors balance capacity and generalization. Together, these choices yield high accuracy at minimal parameter cost, validating that optimal architectural design plus locality-aware attention is a robust strategy for low-resource medical imaging.
\section{Conclusion}

We introduced CoMViT, a compact and generalizable Vision Transformer designed for low-resource medical image classification. By replacing rigid patch-based tokenization with an optimized convolutional tokenizer and carefully selecting lightweight architectural components, CoMViT embeds locality priors and reduces spatial and channel-wise redundancy. The model also removes reliance on a fixed class token by employing learnable sequence pooling and incorporates localized self-attention to focus on discriminative regions.

Extensive evaluation on the MedMNIST2D benchmark demonstrates that CoMViT achieves top-tier accuracy across 12 diverse datasets, outperforming or matching much larger ViTs and CNNs while using significantly fewer parameters. It delivers robust generalization, high interpretability, and excellent scalability—crucial for clinical deployment. Our findings highlight the importance of rethinking tokenization and architectural design in ViTs for medical imaging. CoMViT offers a practical and efficient alternative, advancing the development of lightweight, reliable, and deployable AI models for real-world healthcare settings.

\begin{credits}
\subsubsection{\ackname} This publication has emanated from research conducted with the financial support of Taighde Éireann – Research Ireland under Grant number 18/CRT/6183.  \footnote{For the purpose of Open Access, the author has applied a CC BY public copyright licence to any Author Accepted Manuscript version arising from this submission.}

\subsubsection{\discintname}
The authors have no competing interests to declare that are
relevant to the content of this article.
\end{credits}
%
%
 \sloppy
 \bibliographystyle{splncs04}
 \bibliography{references}

\begin{thebibliography}{10}
\providecommand{\url}[1]{\texttt{#1}}
\providecommand{\urlprefix}{URL }
\providecommand{\doi}[1]{https://doi.org/#1}

\bibitem{aburass_vision_2025}
Aburass, S., Dorgham, O., Al~Shaqsi, J., Abu~Rumman, M., Al-Kadi, O.: Vision {Transformers} in {Medical} {Imaging}: a {Comprehensive} {Review} of {Advancements} and {Applications} {Across} {Multiple} {Diseases}. Journal of Imaging Informatics in Medicine pp. 1--44 (Mar 2025). \doi{10.1007/s10278-025-01481-y}, \url{https://link.springer.com/article/10.1007/s10278-025-01481-y}, publisher: Springer

\bibitem{caron_emerging_2021}
Caron, M., Touvron, H., Misra, I., Jegou, H., Mairal, J., Bojanowski, P., Joulin, A.: Emerging {Properties} in {Self}-{Supervised} {Vision} {Transformers}. In: 2021 {IEEE}/{CVF} {International} {Conference} on {Computer} {Vision} ({ICCV}). pp. 9630--9640. IEEE, Montreal, QC, Canada (Oct 2021). \doi{10.1109/ICCV48922.2021.00951}, \url{https://ieeexplore.ieee.org/document/9709990/}

\bibitem{dosovitskiy_image_2021}
Dosovitskiy, A., Beyer, L., Kolesnikov, A., Weissenborn, D., Zhai, X., Unterthiner, T., Dehghani, M., Minderer, M., Heigold, G., Gelly, S., Uszkoreit, J., Houlsby, N.: An {Image} is {Worth} 16x16 {Words}: {Transformers} for {Image} {Recognition} at {Scale} (Jun 2021), \url{http://arxiv.org/abs/2010.11929}, arXiv:2010.11929 [cs]

\bibitem{du_mdvit_2023}
Du, S., Bayasi, N., Hamarneh, G., Garbi, R.: {MDViT}: {Multi}-domain {Vision} {Transformer} for {Small} {Medical} {Image} {Segmentation} {Datasets} (Oct 2023). \doi{10.1007/978-3-031-43901-8_43}, \url{https://link.springer.com/chapter/10.1007/978-3-031-43901-8_43}

\bibitem{ferdous_spt-swin_2024}
Ferdous, G.J., Sathi, K.A., Hossain, M.A., Dewan, M.A.A.: {SPT}-{Swin}: {A} {Shifted} {Patch} {Tokenization} {Swin} {Transformer} for {Image} {Classification}. IEEE Access  \textbf{12},  117617--117626 (2024). \doi{10.1109/ACCESS.2024.3448304}, \url{https://ieeexplore.ieee.org/document/10643534}

\bibitem{halder_implementing_2024}
Halder, A., Gharami, S., Sadhu, P., Singh, P.K., Woźniak, M., Ijaz, M.F.: Implementing vision transformer for classifying {2D} biomedical images. Scientific Reports  \textbf{14}(1),  12567 (May 2024). \doi{10.1038/s41598-024-63094-9}, \url{https://www.nature.com/articles/s41598-024-63094-9}, publisher: Nature Publishing Group

\bibitem{han_survey_2023}
Han, K., Wang, Y., Chen, H., Chen, X., Guo, J., Liu, Z., Tang, Y., Xiao, A., Xu, C., Xu, Y., Yang, Z., Zhang, Y., Tao, D.: A {Survey} on {Vision} {Transformer}. IEEE Transactions on Pattern Analysis and Machine Intelligence  \textbf{45}(1),  87--110 (Jan 2023). \doi{10.1109/TPAMI.2022.3152247}, \url{https://ieeexplore.ieee.org/abstract/document/9716741}, conference Name: IEEE Transactions on Pattern Analysis and Machine Intelligence

\bibitem{compact}
Hassani, A., Walton, S., Shah, N., Abuduweili, A., Li, J., Shi, H.: Escaping the big data paradigm with compact transformers. CoRR  \textbf{abs/2104.05704} (2021), \url{https://arxiv.org/abs/2104.05704}

\bibitem{hassani_escaping_2022}
Hassani, A., Walton, S., Shah, N., Abuduweili, A., Li, J., Shi, H.: Escaping the {Big} {Data} {Paradigm} with {Compact} {Transformers} (Jun 2022). \doi{10.48550/arXiv.2104.05704}, \url{http://arxiv.org/abs/2104.05704}, arXiv:2104.05704 [cs]

\bibitem{herrmann_cvpr_nodate}
Herrmann, C., Sargent, K., Jiang, L., Zabih, R., Chang, H., Liu, C., Krishnan, D., Sun, D.: {CVPR} 2022 {Open} {Access} {Repository}. \url{https://openaccess.thecvf.com/content/CVPR2022/html/Herrmann_Pyramid_Adversarial_Training_Improves_ViT_Performance_CVPR_2022_paper.html}

\bibitem{lee_improving_2022}
Lee, S., Lee, S., Song, B.C.: Improving {Vision} {Transformers} to {Learn} {Small}-{Size} {Dataset} {From} {Scratch}. IEEE Access  \textbf{10},  123212--123224 (2022). \doi{10.1109/ACCESS.2022.3224044}, \url{https://ieeexplore.ieee.org/document/9957006/}

\bibitem{li_transforming_2023}
Li, J., Chen, J., Tang, Y., Wang, C., Landman, B.A., Zhou, S.K.: Transforming medical imaging with {Transformers}? {A} comparative review of key properties, current progresses, and future perspectives. Medical Image Analysis  \textbf{85},  102762 (Apr 2023). \doi{10.1016/j.media.2023.102762}, \url{https://www.sciencedirect.com/science/article/pii/S1361841523000233}

\bibitem{liu_feature_2022}
Liu, J., Li, Y., Cao, G., Liu, Y., Cao, W.: Feature {Pyramid} {Vision} {Transformer} for {MedMNIST} {Classification} {Decathlon}. In: 2022 {International} {Joint} {Conference} on {Neural} {Networks} ({IJCNN}). pp.~1--8 (Jul 2022). \doi{10.1109/IJCNN55064.2022.9892282}, \url{https://ieeexplore.ieee.org/document/9892282}, iSSN: 2161-4407

\bibitem{manzari_medvit_2023}
Manzari, O.N., Ahmadabadi, H., Kashiani, H., Shokouhi, S.B., Ayatollahi, A.: {MedViT}: {A} {Robust} {Vision} {Transformer} for {Generalized} {Medical} {Image} {Classification}. Computers in Biology and Medicine  \textbf{157},  106791 (May 2023). \doi{10.1016/j.compbiomed.2023.106791}, \url{http://arxiv.org/abs/2302.09462}, arXiv:2302.09462 [cs]

\bibitem{nguyen_semi-supervised_2021}
Nguyen, N.Q., Le, T.S.: A {Semi}-{Supervised} {Learning} {Method} to {Remedy} the {Lack} of {Labeled} {Data}. In: 2021 15th {International} {Conference} on {Advanced} {Computing} and {Applications} ({ACOMP}). pp. 78--84 (Nov 2021). \doi{10.1109/ACOMP53746.2021.00017}, \url{https://ieeexplore.ieee.org/document/9668240}, iSSN: 2688-0202

\bibitem{steiner_how_2022}
Steiner, A., Kolesnikov, A., Zhai, X., Wightman, R., Uszkoreit, J., Beyer, L.: How to train your {ViT}? {Data}, {Augmentation}, and {Regularization} in {Vision} {Transformers} (Jun 2022). \doi{10.48550/arXiv.2106.10270}, \url{http://arxiv.org/abs/2106.10270}, arXiv:2106.10270 [cs]

\bibitem{touvron_training_2021}
Touvron, H., Cord, M., Douze, M., Massa, F., Sablayrolles, A., Jegou, H.: Training data-efficient image transformers \& distillation through attention. In: Proceedings of the 38th {International} {Conference} on {Machine} {Learning}. pp. 10347--10357. PMLR (Jul 2021), \url{https://proceedings.mlr.press/v139/touvron21a.html}, iSSN: 2640-3498

\bibitem{avidan_towards_2022}
Wang, W., Zhang, J., Cao, Y., Shen, Y., Tao, D.: Towards {Data}-{Efficient} {Detection} {Transformers}. In: Avidan, S., Brostow, G., Cissé, M., Farinella, G.M., Hassner, T. (eds.) Computer {Vision} – {ECCV} 2022, vol. 13669, pp. 88--105. Springer Nature Switzerland, Cham (2022). \doi{10.1007/978-3-031-20077-9_6}, \url{https://link.springer.com/10.1007/978-3-031-20077-9_6}, series Title: Lecture Notes in Computer Science

\bibitem{Wightman_PyTorch_Image_Models}
Wightman, R.: {PyTorch Image Models}. \doi{10.5281/zenodo.4414861}, \url{https://github.com/huggingface/pytorch-image-models}

\bibitem{xu_medrdf_2022}
Xu, M., Zhang, T., Zhang, D.: {MedRDF}: {A} {Robust} and {Retrain}-{Less} {Diagnostic} {Framework} for {Medical} {Pretrained} {Models} {Against} {Adversarial} {Attack}. IEEE Transactions on Medical Imaging  \textbf{41}(8),  2130--2143 (Aug 2022). \doi{10.1109/TMI.2022.3156268}, \url{https://ieeexplore.ieee.org/document/9726228}

\bibitem{yang_medmnist_2023}
Yang, J., Shi, R., Wei, D., Liu, Z., Zhao, L., Ke, B., Pfister, H., Ni, B.: {MedMNIST} v2 - {A} large-scale lightweight benchmark for {2D} and {3D} biomedical image classification. Scientific Data  \textbf{10}(1), ~41 (Jan 2023). \doi{10.1038/s41597-022-01721-8}, \url{https://www.nature.com/articles/s41597-022-01721-8}, publisher: Nature Publishing Group

\end{thebibliography}

\end{document}